\documentclass[11pt,a4paper]{article}
\usepackage[hyperref]{acl2020}
\usepackage{times}
\usepackage{latexsym}

\usepackage{microtype}

\aclfinalcopy % Uncomment this line for the final submission
\usepackage{xspace}
\usepackage{graphicx}
\usepackage{listings}
\usepackage{xcolor}
\usepackage{booktabs}
 
\definecolor{codegreen}{rgb}{0,0.6,0}
\definecolor{codegray}{rgb}{0.5,0.5,0.5}
\definecolor{codepurple}{rgb}{0.58,0,0.82}
\definecolor{backcolour}{rgb}{0.95,0.95,0.92}
 
\lstdefinestyle{mystyle}{
    backgroundcolor=\color{backcolour},   
    commentstyle=\color{codegreen},
    keywordstyle=\color{magenta},
    numberstyle=\tiny\color{codegray},
    stringstyle=\color{codepurple},
    basicstyle=\ttfamily\scriptsize,
    breakatwhitespace=false,         
    breaklines=true,                 
    captionpos=b,                    
    keepspaces=true,                 
    numbers=left,                    
    numbersep=5pt,                  
    showspaces=false,                
    showstringspaces=false,
    showtabs=false,                  
    tabsize=2
}
 
\newcommand{\toolkitname}{ConvLab-2\xspace}

\title{\toolkitname: An Open-Source Toolkit for \\
Building, Evaluating, and Diagnosing Dialogue Systems}
\author{Qi Zhu$^\dag$ \quad Zheng Zhang$^\dag$ \quad Yan Fang$^\dag$ \quad Xiang Li$^\dag$ \quad Ryuichi Takanobu$^\dag$   \\ \textbf{Jinchao Li$^\ddag$} \quad \textbf{Baolin Peng$^\ddag$} \quad \textbf{Jianfeng Gao$^\ddag$} \quad \textbf{Xiaoyan Zhu$^\dag$}
\quad \textbf{Minlie Huang$^\dag$\footnotemark[1]}\\
  $^\dag$Dept. of Computer Science and Technology, $^\dag$Institute for Artificial Intelligence, \\
  $^\dag$State Key Lab of Intelligent Technology and Systems,\\
  $^\dag$Beijing National Research Center for Information Science and Technology,\\
  Tsinghua University, Beijing, China\\ $^\ddag$Microsoft Research, Redmond, USA \\
  $^\dag${\small \tt \{zhu-q18,z-zhang15,fangy17,gxly19\}@mails.tsinghua.edu.cn} \\
  $^\ddag${\small \tt \{jincli,bapeng,jfgao\}@microsoft.com} \quad 
  $^\dag${\small \tt \{zxy-dcs,aihuang\}@tsinghua.edu.cn}}

\begin{document}
\maketitle

\renewcommand{\thefootnote}{\fnsymbol{footnote}}
\footnotetext[1]{Corresponding author.}

\renewcommand{\thefootnote}{\arabic{footnote}}

\begin{abstract}
We present \textit{\toolkitname}, an open-source toolkit that enables researchers to build task-oriented dialogue systems with state-of-the-art models, perform an end-to-end evaluation, and diagnose the weakness of systems. 
As the successor of ConvLab \cite{lee-etal-2019-convlab}, \toolkitname inherits ConvLab's framework but integrates more powerful dialogue models and supports more datasets. 
Besides, we have developed an analysis tool and an interactive tool to assist researchers in diagnosing dialogue systems. 
The analysis tool presents rich statistics and summarizes common mistakes from simulated dialogues, which facilitates error analysis and system improvement.
The interactive tool provides a user interface that allows developers to diagnose an assembled dialogue system by interacting with the system and modifying the output of each system component. 
\end{abstract}

\section{Introduction}

\begin{figure}[t]
    \centering
    \includegraphics[width=\linewidth]{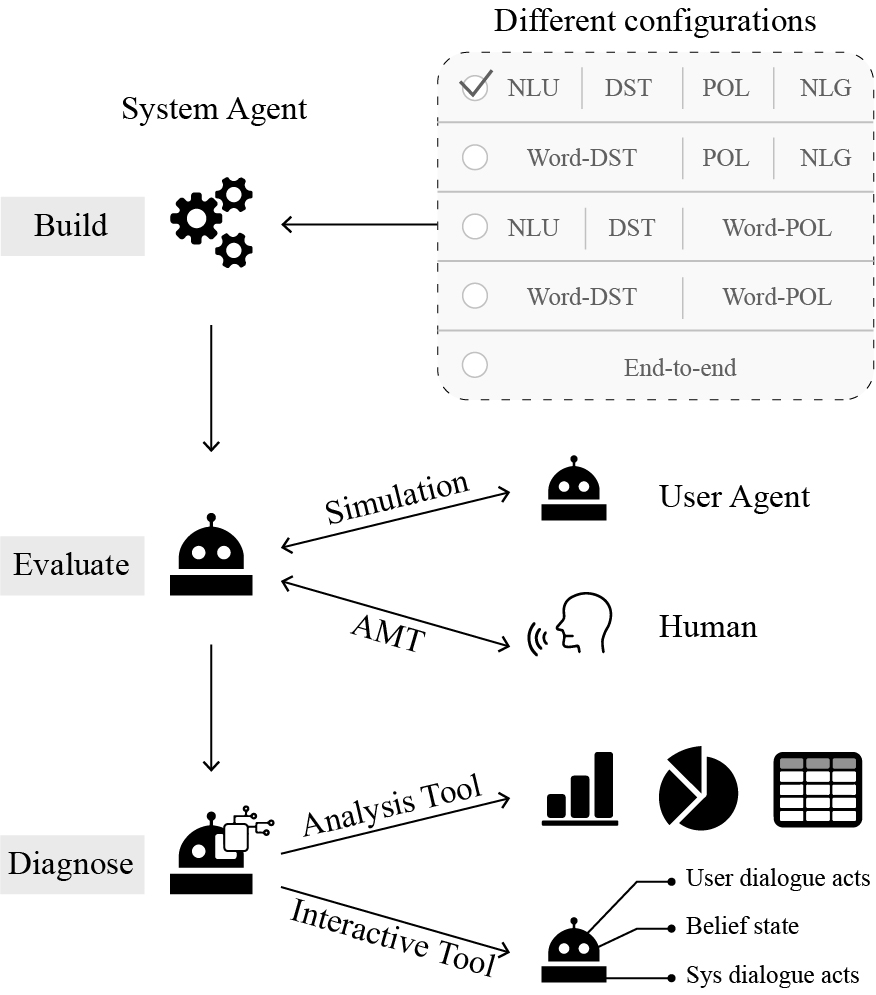}
    \caption{
        Framework of \toolkitname.
        The top block shows different approaches to build a dialogue system.
    }
    \label{fig:framework}
\end{figure}

% toolkit 很重要
% 与以往工作不同，强调与convlab的对比
Task-oriented dialogue systems are gaining increasing
%more and more 
attention in recent years, resulting in a number of datasets \cite{dstc2,Camrest2017,Multiwoz2018,SchemaGuided} and a wide variety of models \cite{wen2015sclstm, HRL, lei2018sequicity, TRADE,gao2019neural}.
However, very few open-source toolkits provide full support to assembling an end-to-end dialogue system with state-of-the-art models, evaluating the performance in an end-to-end fashion, and analyzing the bottleneck both qualitatively and quantitatively.
To fill the gap, we have developed \textbf{\toolkitname} based on our previous dialogue system platform ConvLab \cite{lee-etal-2019-convlab}. 
\toolkitname inherits its predecessor's framework and extend it by integrating many recently proposed state-of-the-art dialogue models. In addition, two powerful tools, namely the analysis tool and the interactive tool, are provided for in-depth error analysis.
\toolkitname will be the development platform for Multi-domain Task-oriented Dialog Challenge II track in the 9th Dialog System Technology Challenge (DSTC9)\footnote{\url{https://sites.google.com/dstc.community/dstc9/home}}.

% pydial parlai rasa plato
% 新的部分，dataset，model
As shown in Figure \ref{fig:framework}, there are many approaches to building a task-oriented dialogue system, ranging from pipeline methods with multiple components to fully end-to-end models.
Previous toolkits %for researchers 
focus on either end-to-end models \cite{miller2017parlai} or one specific component such as dialogue policy (POL) \cite{ultes2017pydial}, while the others toolkits that are designed for developers \cite{bocklisch2017rasa,alex2020plato} do not have state-of-the-art models integrated.
% \cite{ultes2017pydial,miller2017parlai,MDC2018} mainly focus on dialogue policy (POL), while other components such as natural language understanding (NLU) and dialogue state tracking (DST) are less considered.
ConvLab \cite{lee-etal-2019-convlab} is the first toolkit that provides various powerful models for all dialogue components and allows researchers to quickly assemble a complete dialogue system (using a set of recipes).
\toolkitname inherits the flexible framework of ConvLab and imports recently proposed models that achieve state-of-the-art performance. In addition, \toolkitname  supports several large-scale dialogue datasets including CamRest676 \cite{Camrest2017}, MultiWOZ \cite{Multiwoz2018}, DealOrNoDeal \cite{dealornot2017}, and CrossWOZ \cite{zhu2020crosswoz}.

% 新的两个tool
% 放一个 visual 的结果图
To support end-to-end evaluation,  
\toolkitname provides user simulators for automatic evaluation and integrates Amazon Mechanical Turk for human evaluation, similar to ConvLab. 
Moreover, it provides an analysis tool and a human-machine interactive tool for diagnosing a dialogue system. 
Researchers can perform quantitative analysis using the analysis tool.
It presents useful statistics extracted from the conversations between the user simulator and the dialogue system. This information helps reveal the weakness of the system and signifies the direction for further improvement.
With the interactive tool, researchers can perform qualitative analysis by deploying their dialogue systems and conversing with the systems via the webpage. 
During the conversation, the intermediate output of each component in a pipeline system, such as the user dialogue acts and belief state, are presented on the webpage. In this way, the performance of the system can be examined, and the prediction errors of those components can be corrected manually, which helps the developers identify the bottleneck component.
The interactive tool can also be used to collect real-time human-machine dialogues and user feedback for further system improvement.
% Besides qualitative analysis supported by the interactive tool, researchers can also perform quantitative analysis using the analysis tool.
% The analysis tool presents useful statistics extracted from the conversations between the user simulator and the dialogue system, which reveal the weakness of the system and point out the direction of improvement.
% 与ConvLab 相比，提供了更丰富的指标和可视化

% 1 page
\section{\toolkitname}
\subsection{Dialogue Agent}
Each speaker in a conversation is regarded as an agent.
\toolkitname inherits and simplifies ConvLab's framework to accommodate more complicated dialogue agents (e.g., using multiple models for one component) and more general scenarios (e.g., multi-party conversations). 
% As shown at the top of Figure \ref{fig:framework}, the dialogue agent is the core, supported by a task-specific ontology.
Thanks to the flexibility of the agent definition, researchers can build an agent with different types of configurations, such as a traditional pipeline method (as shown in the first layer of the top block in Figure \ref{fig:framework}), a fully end-to-end method (the last layer), and between (other layers) once instantiating corresponding models.
Researchers can also freely customize an agent, such as incorporating two dialogue systems into one agent to cope with multiple tasks.
%Moreover, we treat both dialogue systems and user simulators as agents.
Based on the unified agent definition that both dialogue systems and user simulators are treated as agents,  
\toolkitname supports conversation between two agents and can be extended to more general scenarios involving three or more parties.

% As shown in Figure \ref{fig:framework}, besides automatic evaluation with user simulators and human evaluation using Amazon Mechanical Turk, \toolkitname offers an analysis tool and an interactive tool for further diagnosing.
% As shown in Figure \ref{fig:framework}, \toolkitname offers three methods to evaluate the agent in an end-to-end fashion:
% \begin{itemize}
%     \item Automatic evaluation with the user simulator. Besides basic metrics inherited from ConvLab, the analysis tool collects and presents more information to assist in error analysis.
%     \item Real-time conversation through the interactive tool. The intermediate results of every component of the agent are shown and can be modified manually for detecting the bottleneck.
%     \item Human evaluation via Amazon Mechanical Turk (AMT). \toolkitname adapted the integration of AMT from ConvLab.
% \end{itemize}
% 0.5 page
\subsection{Models}
\toolkitname provides the following models
% \footnote{Please refer to the \toolkitname site for more details.} 
for every possible component in a dialogue agent. Note that compared to ConvLab, newly integrated models in \toolkitname are marked in bold.
Researchers can easily add their models by implementing the interface of the corresponding component.
We will keep adding state-of-the-art models to reflect the latest progress in task-oriented dialogue.

\subsubsection{Natural Language Understanding}
The natural language understanding (NLU) component, which is used to parse the other agent's intent, takes an utterance as input and outputs the corresponding dialogue acts.
\toolkitname provides three models: Semantic Tuple Classifier (STC) \cite{mairesse2009spoken}, MILU \cite{lee-etal-2019-convlab}, and \textbf{BERTNLU}. 
BERT \cite{bert2019} has shown strong performance in many NLP tasks. 
Thus, \toolkitname proposes a new BERTNLU model.
BERTNLU adds two MLPs on top of BERT for intent classification and slot tagging, respectively, and fine-tunes all parameters on the specified tasks. 
BERTNLU achieves the best performance on MultiWOZ in comparison with other models.

\subsubsection{Dialogue State Tracking}
The dialogue state tracking (DST) component updates the belief state, which contains the constraints and requirements of the other agent (such as a user). \toolkitname provides a rule-based tracker that takes dialogue acts parsed by the NLU as input.

\subsubsection{Word-level Dialogue State Tracking}
Word-level DST obtains the belief state directly from the dialogue history. \toolkitname integrates four models: MDBT \cite{ramadan2018large}, \textbf{SUMBT} \cite{lee2019sumbt}, and \textbf{TRADE} \cite{TRADE}.
TRADE generates the belief state from utterances using a copy mechanism and achieves state-of-the-art performance on MultiWOZ.

\subsubsection{Dialogue Policy}
Dialogue policy receives the belief state and outputs system dialogue acts.
\toolkitname provides a rule-based policy, 
a simple neural policy that learns directly from the corpus using imitation learning,
and reinforcement learning policies including REINFORCE \cite{williams1992simple}, PPO \cite{schulman2017proximal}, and \textbf{GDPL} \cite{takanobu2019guided}.
GDPL achieves state-of-the-art performance on MultiWOZ.

\subsubsection{Natural Language Generation}
The natural language generation (NLG) component transforms dialogue acts into a natural language sentence. \toolkitname provides a template-based method and SC-LSTM \cite{wen2015sclstm}.

\subsubsection{Word-level Policy}
Word-level policy directly generates a natural language response (rather than dialogue acts) according to the dialogue history and the belief state.
\toolkitname integrates three models: MDRG \cite{MDRG}, \textbf{HDSA} \cite{chen2019semantically}, and \textbf{LaRL} \cite{zhao2019rethinking}. MDRG is the baseline model proposed by \citet{Multiwoz2018} on MultiWOZ, while HDSA and LaRL achieve much stronger performance on this dataset.

\subsubsection{User Policy}
User policy is the core of a user simulator. It takes a pre-set user goal and system dialogue acts as input and outputs user dialogue acts.
\toolkitname provides an agenda-based \cite{agenda2007schatzmann} model and neural network-based models including HUS and its variational variants \cite{gur2018user}. To perform end-to-end simulation, researchers can equip the user policy with NLU and NLG components to assemble a complete user simulator.

\subsubsection{End-to-end Model}
A fully end-to-end dialogue model receives the dialogue history and generates a response in natural language directly.
\toolkitname extends Sequicity \cite{lei2018sequicity} to multi-domain scenarios: when the model senses that the current domain has switched, it resets the belief span, which records information of the current domain.
\toolkitname also integrates \textbf{DAMD} \cite{Zhang2019damd} which obtains state-of-the-art results on MultiWOZ.
% zhangzheng
As for the DealOrNoDeal dataset, we provide the \textbf{ROLLOUTS RL} policy
proposed by \citet{dealornot2017}.
% , which is based on the REINFORCE algorithm \cite{williams1992simple} and uses Monte Carlo tree search to choose the best action to respond that maximizes the reward.

% 1 page
\subsection{Datasets}
Compared with ConvLab, \toolkitname can integrate a new dataset more conveniently.
For each dataset, \toolkitname provides a unified data loader that can be used by all the models, thus separating data processing from the model definition.
Currently, \toolkitname supports four task-oriented dialogue datasets, including CamRest676 \cite{Camrest2017}, MultiWOZ \cite{eric2019multiwoz21}, DealOrNoDeal \cite{dealornot2017}, and CrossWOZ \cite{zhu2020crosswoz}.

\subsubsection{CamRest676}
CamRest676 \cite{Camrest2017} is a Wizard-of-Oz dataset,  
consisting of 676 dialogues in a restaurant domain.
%, with dialogue acts at both user and system sides.
% The corresponding ontology and a knowledge base are also provided.
\toolkitname offers an agenda-based user simulator and a complete set of models for building a traditional pipeline dialogue system on the CamRest676 dataset.

\subsubsection{MultiWOZ}
MultiWOZ \cite{Multiwoz2018} is a large-scale multi-domain Wizard-of-Oz dataset.
%that has attracted much attention recently.
%The conversations span over seven domains, including Attraction, Restaurant, Hotel, Taxi, Train, Hospital, and Police.
It consists of 10,438 dialogues with system dialogue acts and belief states.
However, user dialogue acts are missing, and belief state annotations and dialogue utterances are noisy.
To address these issues, Convlab \cite{lee-etal-2019-convlab} annotated user dialogue acts automatically using heuristics. \citet{eric2019multiwoz21} further re-annotated the belief states and utterances, resulting in the MultiWOZ 2.1 dataset.

\subsubsection{DealOrNoDeal}
DealOrNoDeal \cite{dealornot2017} is a dataset of human-human negotiations on a multi-issue bargaining task.
It contains 5,805 dialogues based on 2,236 unique scenarios. %The agents cannot observe each other's value function and should reach an agreement through conversation.
On this dataset, \toolkitname implements ROLLOUTS RL \cite{dealornot2017} and LaRL \cite{zhao2019rethinking} models.

\begin{figure*}[t]
    \begin{minipage}{0.52\linewidth}
    \includegraphics[width=\linewidth]{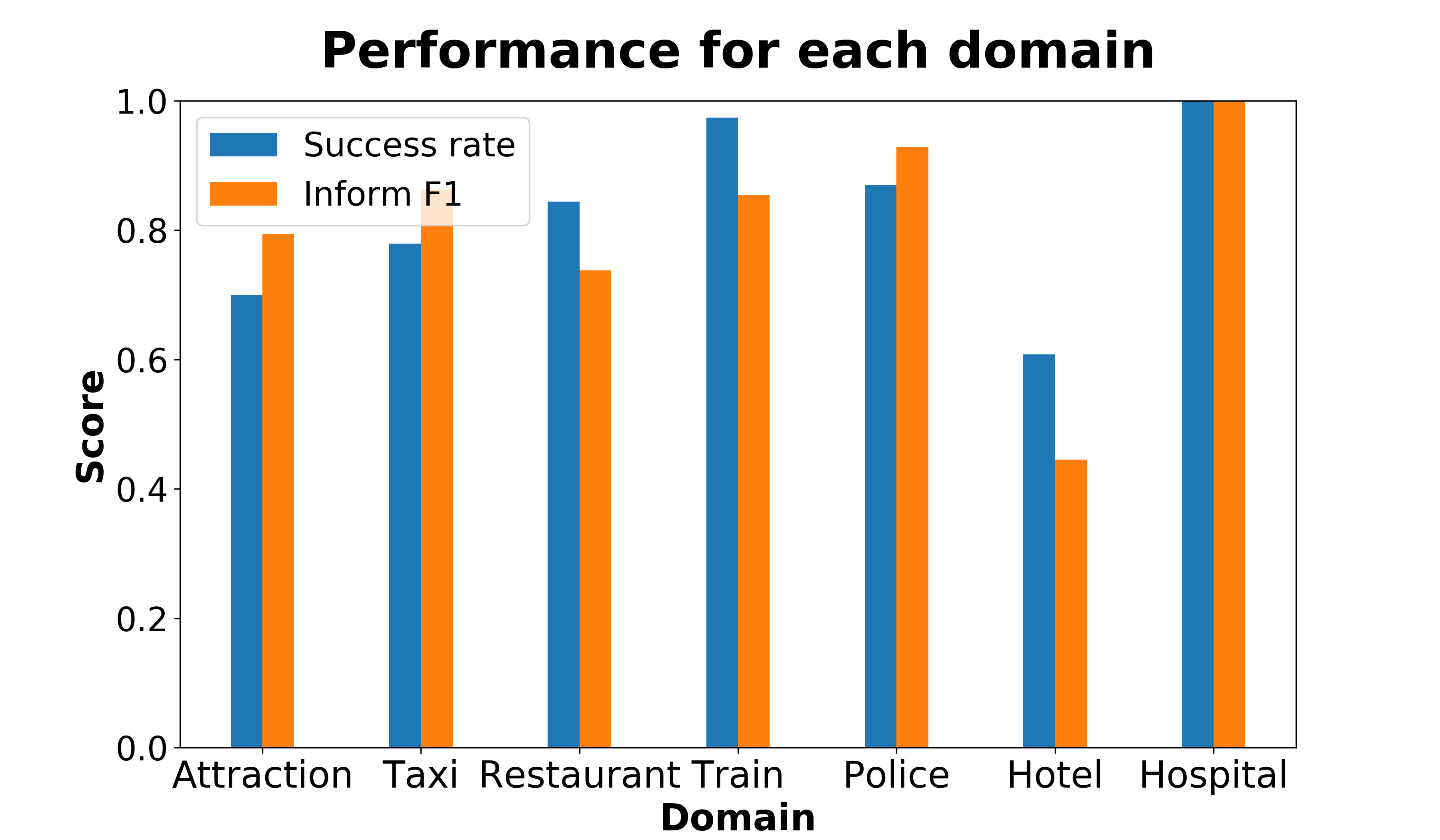}
    \end{minipage}
    \begin{minipage}{0.46\linewidth}
    \includegraphics[width=\linewidth]{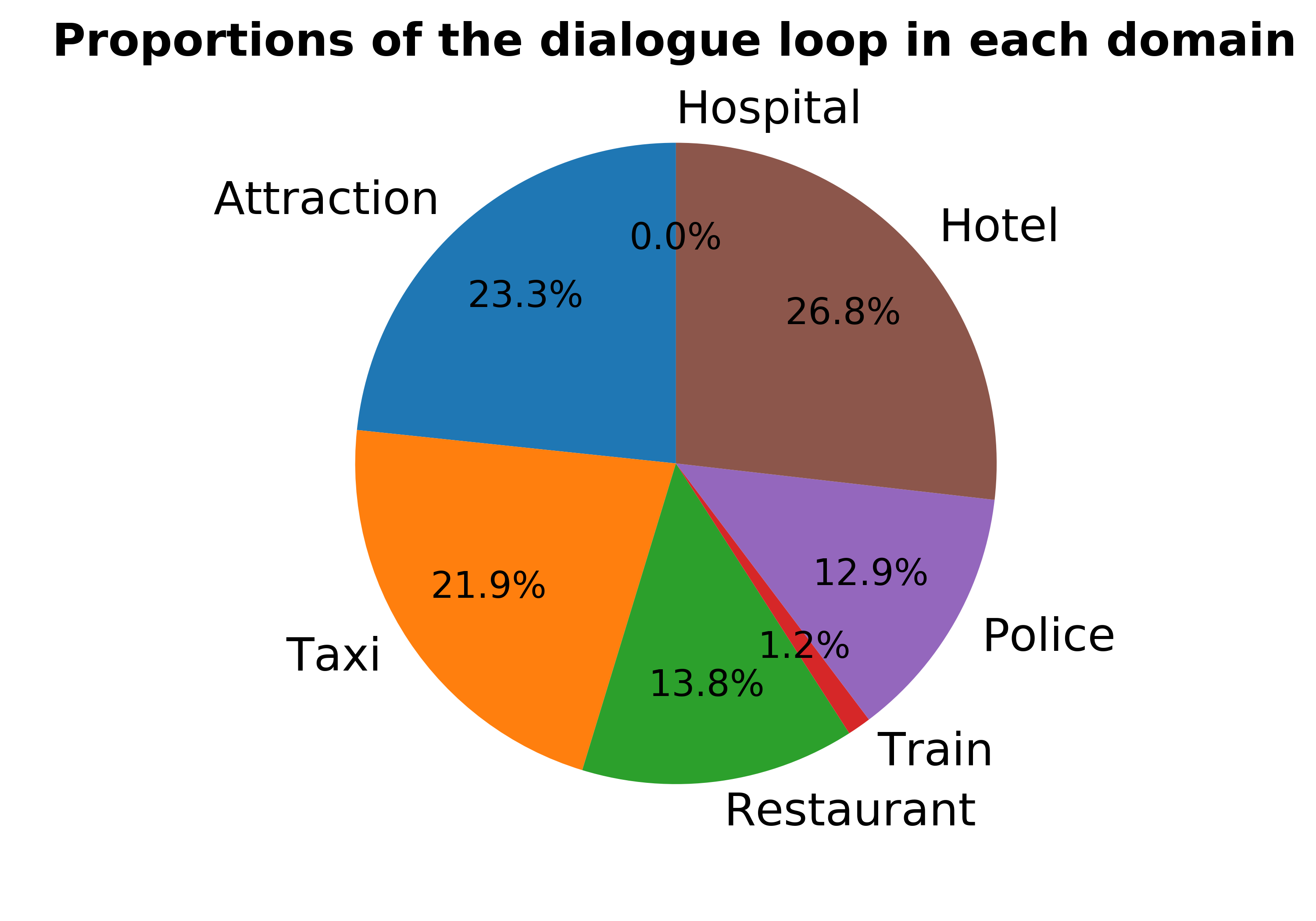}
    \end{minipage}
    \caption{Performance of the demo system in Section \ref{sec:demo}. \textbf{Left:} Success rate and inform F1 for each domain. \textbf{Right:} Proportions of the dialogue loop in each domain.}
    \label{fig:analysis_res}
\end{figure*}

\subsubsection{CrossWOZ}
CrossWOZ \cite{zhu2020crosswoz} is the first large-scale Chinese multi-domain Wizard-of-Oz dataset proposed recently.
It contains 6,012 dialogues spanning over five domains.
% About 60\% of the user goals have inter-domain dependencies that imply transitions across domains.
Besides dialogue acts and belief states, the annotations of user states, which indicate the completion of a user goal, are also provided.
\toolkitname offers a rule-based user simulator and a complete set of models for building a pipeline system on the CrossWOZ dataset.

\begin{table}[]
    \centering
    \begin{tabular}{p{7.3cm}}
        \toprule
        \textbf{Overall results:} \\
        Success Rate: 60.8\%; inform F1: 44.5\% \\
        \midrule
        \textbf{Most confusing user dialogue acts:} \\
        Request-Hotel-Post-?\\%: 191 times\\
        \quad - 34\%: Request-Hospital-Post-?\\
        \quad - 32\%: Request-Attraction-Post-?\\
        % Inform-Hotel-Parking-yes: 295 times\\
        % - 17\%: Inform-Hotel-Internet-yes \\
        % - 14\%: Inform-Hotel-Type-guesthouse \\
        Request-Hotel-Addr-?\\%: 269 times\\
        \quad - 29\%: Request-Attraction-Addr-? \\
        \quad - 28\%: Request-Restaurant-Addr-? \\
        % Inform-Hotel-Internet-yes: 264 times\\
        % - 18\%: Inform-Hotel-Parking-yes\\
        % - 13\%: Inform-Hotel-Type-guesthouse\\
        Request-Hotel-Phone-?\\%: 254 times\\
        \quad - 26\%: Request-Restaurant-Phone-?\\
        \quad - 26\%: Request-Attraction-Phone-?\\
        \midrule
        \textbf{Invalid system dialogue acts:}\\
        \quad - 31\%: Inform-Hotel-Parking\\
        \quad - 28\%: Inform-Hotel-Internet\\
        \textbf{Redundant system dialogue acts:}\\
        \quad - 34\%: Inform-Hotel-Stars\\
        \textbf{Missing system dialogue acts:}\\
        \quad - 25\%: Inform-Hotel-Phone\\
        \midrule
        \textbf{Most confusing system dialogue acts:}\\
        Recommend-Hotel-Parking-yes\\
        \quad - 21\%: Recommend-Hotel-Parking-none\\
        \quad - 18\%: Inform-Hotel-Parking-none\\
        Inform-Hotel-Parking-yes\\
        \quad - 17\%: Inform-Hotel-Parking-none\\
        Inform-Hotel-Stars-4\\
        \quad - 16\%: Inform-Hotel-Internet-none\\
        % Inform-Hotel-Internet-yes\\
        % \quad - 15\%: Inform-Hotel-Internet-none\\
        \midrule
        \textbf{User dialogue acts that cause loop:}\\
        \quad - 53\% Request-Hotel-Phone-?\\
        \quad - 21\% Request-Hotel-Post-?\\
        \quad - 14\% Request-Hotel-Addr-?\\
        \bottomrule
    \end{tabular}
    \caption{
    Comprehensive results (partial) of the demo system in Section \ref{sec:demo} for the Hotel domain.
    To save space, only the most frequent errors are presented.
    }
    \label{tab:analysis_res}
\end{table}

% 0.5 page
\subsection{Analysis Tool}
To evaluate a dialogue system quantitatively, \toolkitname offers an analysis tool to perform an end-to-end evaluation with a specified user simulator and generate an HTML report which contains rich statistics of simulated dialogues.
Charts and tables are used in the test report for better demonstration.
Partial results of a demo system in Section \ref{sec:demo} are shown in Figure \ref{fig:analysis_res} and Table \ref{tab:analysis_res}.
\begin{figure*}[h]
    \centering
    \includegraphics[width=0.8\linewidth]{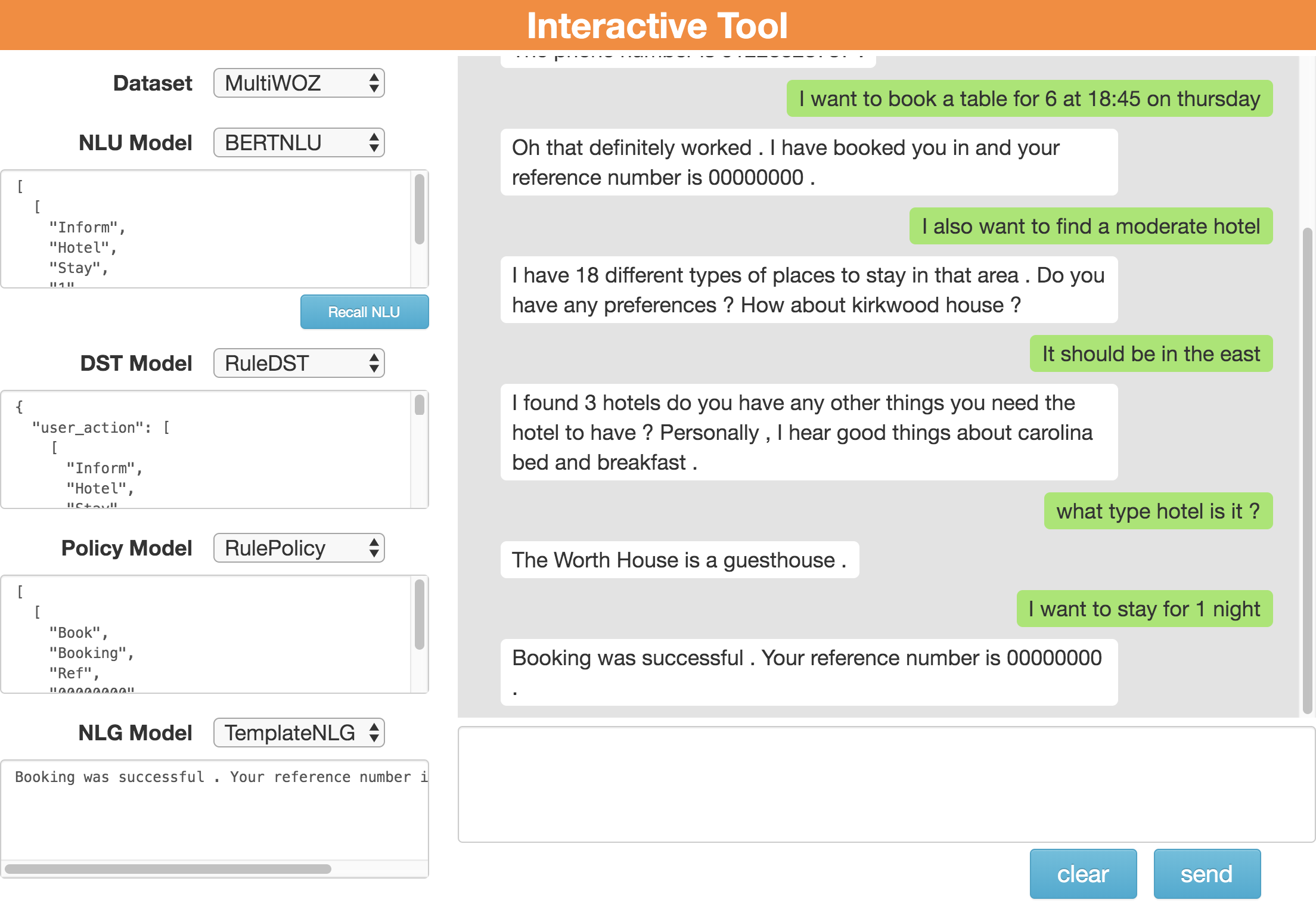}
    \caption{
        The interface of the Interactive Tool. %The reference number in this example is fake.
    }
    \label{fig:interactive_tool}
\end{figure*}
Currently, the report contains the following pieces of information for each task domain:
\begin{itemize}
    \item Metrics for overall performance such as success rate, inform F1, average turn number, etc.
    \item Common errors of the NLU component, such as the confusion matrix of user dialogue acts. For the example in Table \ref{tab:analysis_res}, 34\% of the requests for the Postcode in the Hotel domain are misinterpreted as the requests in the Hospital domain.
    \item Frequent invalid, redundant, and missing system dialogue acts predicted by the dialogue policy.
    \item The system dialogue acts from which the NLG component generates responses that confuse the user simulator. For the example in Table \ref{tab:analysis_res}, it is hard to inform the user that the hotel has free parking.
    \item The causes of dialogue loops. Dialogue loop is the situation that the user keeps repeating the same request until the max turn number is reached.
    % This situation is because the system fails to provide the required information.
    This result shows the requests that are hard for the system to handle.
    % This is because the system fails to provide the required information.
    % The starting points of dialogue loops, from where the user repeats the same request because the system fails to provide the required information.
\end{itemize}

%Besides the individual system evaluation, 
The analysis tool also supports the comparison between different dialogue systems that interact with the same user simulator.
The above statistics and comparison results can significantly facilitate error analysis and system improvement.
% We plan to provide a more comprehensive evaluation in the future.
%keep adding useful information to evaluate the system comprehensively.
% 0.5 page

% \begin{figure*}[t]
%     \centering
%     \includegraphics[width=0.9\linewidth]{img/interactive_tool.png}
%     \caption{
%         Interface of the Interactive Tool. %The reference number in this example is fake.
%     }
%     \label{fig:interactive_tool}
% \end{figure*}

% 0.5 page

\subsection{Interactive Tool}

\toolkitname provides an interactive tool that enables researchers to converse with a dialogue system through a graphical user interface and modify intermediate results to correct system errors.
% \footnote{Demo video: https://drive.google.com/file/d/1HR3mjhgLL0g9IbqU443NsH2G0-PpAsog/view}

As shown in Figure \ref{fig:interactive_tool}, researchers can customize their dialogue system by selecting the dataset and the model of each component. Then, they can interact with the system via the user interface. During a conversation, the output of each component is displayed on the left side as a JSON formatted string, including the user dialogue acts parsed by the NLU, the belief state tracked by the DST, the system dialogue acts selected by the policy and the final system response generated by the NLG.
By showing both the dialogue history and the component outputs, the researchers can get a good understanding of how their system works.

In addition to the fine-grained system output, the interactive tool also supports intermediate output modification. When a component makes a mistake and the dialogue fails to continue, the researchers can correct the JSON output of that component to redirect the conversation by replacing the original output with the correct one. This function is helpful when the researchers are debugging a specific component.

In consideration of the compatibility across platforms, the interactive tool is deployed as a web service that can be accessed via a web browser. To use self-defined models, the researchers have to edit a configuration file, which defines all available models for each component. The researchers can also add their own models into the configuration file easily.

\section{Demo}
\label{sec:demo}
This section demonstrates how to use \toolkitname to build, evaluate, and diagnose a traditional pipeline dialogue system developed on the MultiWOZ dataset.

\begin{lstlisting}[language=Python, caption={Example code for the demo.}]
import ... # import necessary modules
# Create models for each component
# Parameters are omitted for simplicity
sys_nlu = BERTNLU(...)
sys_dst = RuleDST(...)
sys_policy = RulePolicy(...)
sys_nlg = TemplateNLG(...)
# Assemble a pipeline system named "sys"
sys_agent = PipelineAgent(sys_nlu, sys_dst, sys_policy, sys_nlg, name="sys")
# Build a user simulator similarly but without DST
user_nlu = BERTNLU(...)
user_policy = RulePolicy(...)
user_nlg = TemplateNLG(...)
user_agent = PipelineAgent(user_nlu, None, user_policy, user_nlg, name="user")
# Create an evaluator and a conversation environment
evaluator = MultiWozEvaluator()
sess = BiSession(sys_agent, user_agent, evaluator)
# Start simulation
sess.init_session()
sys_utt = ""
while True:
    sys_utt, user_utt, sess_over, reward = sess.next_turn(sys_utt)
    if sess_over:
        break
print(sess.evaluator.task_success())
print(sess.evaluator.inform_F1())
# Use the analysis tool to generate a test report
analyzer = Analyzer(user_agent, dataset="MultiWOZ")
analyzer.comprehensive_analyze(sys_agent, total_dialog=1000)
# Compare multiple systems
sys_agent2 = PipelineAgent(MILU(...), sys_dst, sys_policy, sys_nlg, name="sys")
analyzer.compare_models(agent_list=[sys_agent, sys_agent2], model_name=["bertnlu", "milu"], total_dialog=1000)
\end{lstlisting}

To build such a dialogue system, we need to instantiate a model for each component and assemble them into a complete agent. 
As shown in the above code, the system consists of a BERTNLU, a rule-based DST, a rule-based system policy, and a template-based NLG.
Likewise, we can build a user simulator that consists of a BERTNLU, an agenda-based user policy, and a template-based NLG.
Thanks to the flexibility of the framework, the DST of the simulator can be \texttt{None}, which means passing the parsed dialogue acts directly to the policy without the belief state.

For end-to-end evaluation, \toolkitname provides a \texttt{BiSession} class, which takes a system, a simulator, and an evaluator as inputs.
Then this class can be used to simulate dialogues and calculate end-to-end evaluation metrics.
For example, the task success rate of the system is 64.2\%, and the inform F1 is 67.0\% for 1000 simulated dialogues.
In addition to automatic evaluation, \toolkitname can perform human evaluation via Amazon Mechanical Turk using the same system agent.

Then the analysis tool can be used to perform a comprehensive evaluation.
Equipped with a user simulator, the tool can analyze and compare multiple systems.
Some results are shown in Figure \ref{fig:analysis_res} and Table \ref{tab:analysis_res}. We collected statistics from 1000 simulated dialogues and found that
\begin{itemize}
    \item The demo system performs the poorest in the Hotel domain but always completes the goal in the Hospital domain.
    \item The sub-task in the Hotel domain is more likely to cause dialogue loops than in other domains. More than half of the loops in the Hotel domain are caused by the user request for the phone number.
    \item One of the most common errors of the NLU component is misinterpreting the domain of user dialogue acts.
    For example, the user request for the Postcode, address, and phone number in the Hotel domain is often parsed as in other domains.
    \item In the Hotel domain, the dialogue acts whose slots are \texttt{Parking} are much harder to be perceived than other dialogue acts.
\end{itemize}

The researchers can further diagnose their system by observing fine-grained output and rescuing a failed dialogue using our provided interactive tool.
An example is shown in Figure \ref{fig:interactive_tool}, in which at first the BERTNLU falsely identified the domain as \textit{Restaurant}.
After correcting the domain to \textit{Hotel} manually, a \textit{Recall NLU} button appears. 
By clicking the button, the dialogue system reruns this turn by skipping the NLU module and directly use the corrected NLU output.
Combined with the observations from the analysis tool, alleviating the domain confusion problem of the NLU component may significantly improve system performance.

% 1 page

\section{Code and Resources}

\toolkitname is publicly available on \url{https://github.com/thu-coai/ConvLab-2}.
Resources such as datasets, trained models, tutorials, and demo video are also released.
We will keep track of new datasets and state-of-the-art models.
Contributions from the community are always welcome.

\section{Conclusion}
We present \toolkitname, an open-source toolkit for building, evaluating, and diagnosing a task-oriented dialogue system.
Based on ConvLab \cite{lee-etal-2019-convlab}, \toolkitname integrates more powerful models, supports more datasets, and develops an analysis tool and an interactive tool for comprehensive end-to-end evaluation.
For demonstration, we give an example of using \toolkitname to build, evaluate, and diagnose a system on the MultiWOZ dataset.
We hope that \toolkitname is instrumental in promoting the research on task-oriented dialogue.
% and also benefit from the latest progress.

\section*{Acknowledgments}
This work was jointly supported by the NSFC projects (Key project with No. 61936010 and regular project with No. 61876096), and
the National Key R\&D Program of China (Grant No. 2018YFC0830200). We thank THUNUS NExT Joint-Lab for the support.

\newpage

\bibliography{acl2020}
\bibliographystyle{acl_natbib}

\end{document}